\documentclass[10pt,onecolumn]{IEEEtran_arxiv}

\usepackage{cite}
\usepackage[colorlinks=true, citecolor=blue, linkcolor=blue, urlcolor=blue]{hyperref}

\usepackage{comment}

\ifCLASSINFOpdf
   \usepackage[pdftex]{graphicx}
   \graphicspath{{figs/}}
   \DeclareGraphicsExtensions{.pdf,.jpeg,.png}
\else
   \usepackage[dvips]{graphicx}
   \graphicspath{{../figs/}}
   \DeclareGraphicsExtensions{.eps}
\fi

\usepackage{float}

\usepackage[cmex10]{amsmath}
\usepackage{amsthm}
\usepackage{amssymb}
\usepackage{booktabs}

\usepackage{tikz}
\usetikzlibrary{shapes.geometric, arrows, positioning, fit, calc}

\usepackage{array}

\ifCLASSOPTIONcompsoc
  \usepackage[caption=false,font=normalsize,labelfont=sf,textfont=sf]{subfig}
\else
  \usepackage[caption=false,font=footnotesize]{subfig}
\fi

\usepackage{url}

\usepackage{orcidlink}

\newif\iffinal
\finaltrue
\newcommand{\cmtid}{142}

\iffinal
\else
\usepackage[switch]{lineno}
\fi

\begin{document}

\title{Phase-Preserving Trimodal Transformer for Tropical Forest Biomass Estimation Using Optical and PolInSAR Data\\[1ex]
\Large PREPRINT}

\iffinal
\author{
    Luiz Felipe Parente Santiago\orcidlink{0009-0005-6864-2450}$^{1, 2}$, 
    Felipe Ferrari\orcidlink{0000-0003-0705-9593}$^3$, 
    Daniel Rodrigues dos Santos\orcidlink{0000-0001-7977-7426}$^3$ and 
    Rosiane de Freitas\orcidlink{0000-0002-7608-2052}$^1$\\
    \vspace{0.15cm}
     $^1$Instituto de Computação, Universidade Federal do Amazonas (IComp/UFAM), Manaus-AM, Brazil\\
    \vspace{0.15cm}
    $^2$Instituto de Pesquisas do Exército na Amazônia (IPEAM), Manaus-AM, Brazil\\
    \vspace{0.15cm}
    $^3$Instituto Militar de Engenharia (IME), Rio de Janeiro-RJ, Brazil\\
    \vspace{0.15cm}
    Email: \{luizparente, rosiane\}@icomp.ufam.edu.br, \{daniel.rodrigues, ferrari\}@ime.eb.br
}
\else
  \author{SIBGRAPI Paper ID: \cmtid \\ }
  \linenumbers
\fi

\maketitle

\begin{abstract}
The accurate estimation of Above-Ground Biomass (AGB) in mature tropical forests remains a critical challenge in remote sensing, primarily due to the saturation of Synthetic Aperture Radar (SAR) signals in high-density areas and persistent cloud cover affecting optical imagery. To overcome these physical limitations, we propose the Trimodal Coherent Co-attention Transformer (TCCT), a physics-informed deep learning architecture. The TCCT natively fuses optical surface reflectance (Landsat-5) with complex-valued Polarimetric SAR Interferometry (PolInSAR) data from both P and L bands. Unlike traditional fusion methods, our architecture employs complex-valued encoders to preserve spatial phase coherence, coupled with a dynamic co-attention mechanism that acts as an adaptive gating module, reducing the weight of cloud-corrupted optical pixels and shifting reliance to microwave phase data. We also derived a localized spatial allometric calibration model via Levenberg-Marquardt optimization, tailored to the specific wood density of the Paracou region in the Amazon basin. Evaluated using a two-stage protocol, the TCCT first underwent a rigorous 5-fold cross-validation to establish robust global weights (achieving a global RMSE of 4.19 m). Subsequently, following a localized spatial fine-tuning phase over 200 epochs, the model attained an absolute RMSE of 3.78 m and an $R^2$ of 0.33 for Canopy Height Models (CHM), outperforming standard Random Forest, CNN, and Vision Transformer baselines. Our ablation study confirms that preserving phase coherence mitigates deep-canopy signal saturation. When converted to AGB, the fine-tuned TCCT map yielded a Relative RMSE (rRMSE) of 4.51\% in dense forest areas above 50 Mg/ha. By meeting the European Space Agency (ESA) BIOMASS mission requirement of less than 20\% error, the TCCT provides a robust framework for continuous carbon stock mapping in tropical biomes.
\end{abstract}

\IEEEpeerreviewmaketitle

\section{Introduction}\label{sec:Introduction}

The Amazon Rainforest plays a critical role in the global carbon cycle and climate regulation. The necessity for precise valuation of its resources was consolidated during the 30th United Nations Conference of the Parties on Climate Change (COP-30), which established strict accuracy targets for tracking forest carbon stocks to validate international carbon credit markets (REDD+) \cite{saatchi2011, asner2014}. Consequently, promoting emissions reduction is now inherently tied to reliable, data-driven monitoring of forest degradation \cite{kuck2021}.

In this context, the accurate estimation of Above-Ground Biomass (AGB) is the primary metric for mapping forest carbon sequestration. However, acquiring precise continuous measurements over vast tropical biomes remains highly challenging. Optical sensors provide rich multispectral reflectance data but are severely limited by persistent cloud cover. Conversely, SAR sensors (such as P and L bands) penetrate clouds and interact directly with the forest canopy and trunks \cite{cheng2023, ref_sibgrapi_foresteyes}. Despite this advantage, traditional SAR approaches face a critical physical barrier: signal saturation. In high-density primary forests, such as the Paracou region, AGB frequently exceeds 300 Mg/ha. In these scenarios, the radar backscatter signal typically saturates between 100 and 150 Mg/ha, leading to underestimations of giant trees and massive error margins when generic pantropical allometric equations \cite{chave2014} are applied.

\begin{figure}[!htbp]
\centerline{\includegraphics[width=\columnwidth]{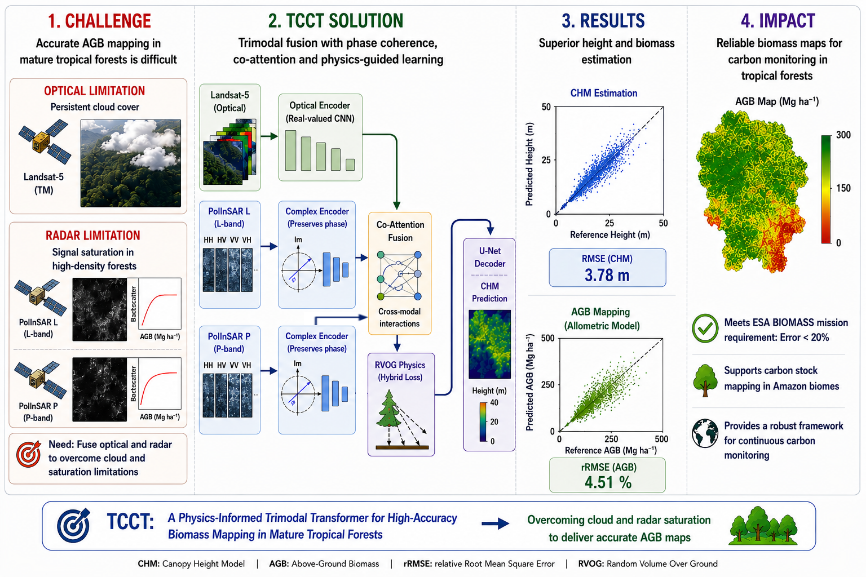}}
\caption{Graphical Abstract: The proposed pipeline integrates trimodal remote sensing data, leveraging phase coherence and co-attention within a U-Net decoder framework to yield Canopy Height Models (CHM) and Above-Ground Biomass (AGB) maps.}
\label{fig:graph_abstract}
\end{figure}

To overcome these limitations, we propose a physics-informed deep learning architecture named the Trimodal Coherent Co-attention Transformer (TCCT). As summarized in the graphical abstract (Fig. \ref{fig:graph_abstract}), the TCCT natively fuses optical surface reflectance with complex-valued PolInSAR data. Inspired by advances in attention mechanisms \cite{ref_sibgrapi_attention1, ref_sibgrapi_attention2}, the TCCT leverages a dynamic co-attention module that shifts reliance between sensors pixel-by-pixel, effectively bypassing cloud corruption and mitigating radar saturation. Furthermore, we address the biological modeling gap by deriving a localized spatial allometric calibration specifically for the Paracou biome.

The main contributions of this paper are summarized as follows:
\begin{itemize}
    \item A multimodal architecture (TCCT) that utilizes complex-valued encoders to preserve spatial phase coherence ($\Delta\phi$) alongside optical data.
    \item A Transformer-based co-attention mechanism capable of dynamically weighting the reliability of Landsat-5, SAR L-Band, and SAR P-Band inputs in real-time.
    \item A customized spatial allometric calibration ($AGB = 162.17 \times H^{0.2486}$) that models the specific wood density of the Amazonian Paracou region.
    \item Extensive validation via a two-stage protocol (5-fold cross-validation and localized fine-tuning) demonstrating that the TCCT achieves a final absolute CHM Root Mean Square Error (RMSE) of 3.78 m. Applied to dense forests (> 50 Mg/ha), the derived AGB map achieved a relative RMSE (rRMSE) of 4.51\%, fulfilling the 20\% error barrier demanded by the ESA BIOMASS mission \cite{letoan2011}.
\end{itemize}

The remainder of this paper is organized as follows. Section 2 reviews related works. Section 3 details the TCCT architecture. Section 4 presents the experimental setup, spatial allometric calibration, and mapping results. Section 5 concludes the paper.

\section{Related Works}\label{sec:RelatedWorks}

The estimation of forest biomass has increasingly relied on the intersection of remote sensing and deep learning. This section reviews advancements in multi-sensor data fusion and specialized architectures for radar processing.

\subsection{Remote Sensing and Multi-Sensor Fusion}
Optical imagery has been extensively applied to forest biomass estimation and degradation monitoring \cite{cheng2023, kuck2021}. Amazonas et al. \cite{ref_sibgrapi_datasets} demonstrated the efficacy of high-resolution satellite imagery for generating forest datasets. However, these optical modalities are inherently susceptible to atmospheric interference. To mitigate this, the computer vision community has investigated the fusion of optical and SAR data \cite{hong2021}. Queiroz et al. \cite{ref_sibgrapi_foresteyes} evaluated the performance of combining optical and SAR imagery for superpixel-based deforestation mapping, demonstrating that multi-sensor approaches significantly improve classification robustness. Similarly, the exploration of spatio-temporal novelty detection and cross-domain aerial monitoring \cite{ref_sibgrapi_segmentation, ref_sibgrapi_synthetic} underscores the necessity for adaptable models. Nevertheless, standard early-fusion techniques still struggle to dynamically penalize corrupted optical inputs in dense tropical applications.

\subsection{Deep Learning for SAR Processing}
The unique statistical properties of SAR backscatter demand specialized neural network architectures. Goldberg and Neto \cite{ref_sibgrapi_convolutional} proposed a parameter estimation-inspired convolutional block specifically tailored for SAR data. While such innovations enhance amplitude-based SAR processing, they generally operate in the real-number domain, discarding the spatial phase coherence ($\Delta\phi$) available in PolInSAR. Preserving this phase information is critical for estimating tree height in dense forests where backscatter amplitude natively saturates.

\subsection{Attention Mechanisms and Transformers}
The advent of attention mechanisms has revolutionized feature extraction. Within the SIBGRAPI community, Lopez-Cabrejos et al. \cite{ref_sibgrapi_attention1} integrated CNNs with attention mechanisms to suppress distortions in image reconstruction tasks, while Filho et al. \cite{ref_sibgrapi_ViT} demonstrated the power of Vision Transformers (ViT) using hierarchical shifted windows. Architectures utilizing adaptive attention to selectively weight hierarchical features \cite{ref_sibgrapi_attention2} offer a compelling paradigm that can be adapted for remote sensing.

\subsection{Addressing the Literature Gap}
To the best of our knowledge, no previous study has simultaneously integrated the radiometric richness of optical data with the complex-valued phase information of PolInSAR while dynamically resolving sensor conflicts via a U-Net decoding scheme. Our proposed TCCT addresses these limitations by extending the adaptive attention paradigm \cite{ref_sibgrapi_attention2} into a multimodal co-attention mechanism. 

\section{Methodology}\label{sec:Methodology}

The proposed TCCT architecture handles the multi-sensor alignment and physical limitations inherent in tropical forest biomass estimation. This section delineates the mathematical formalization of the complex-valued encoders, the co-attention alignment mechanism, and the hybrid optimization framework.

\subsection{Architectural Overview and Data Pre-processing}
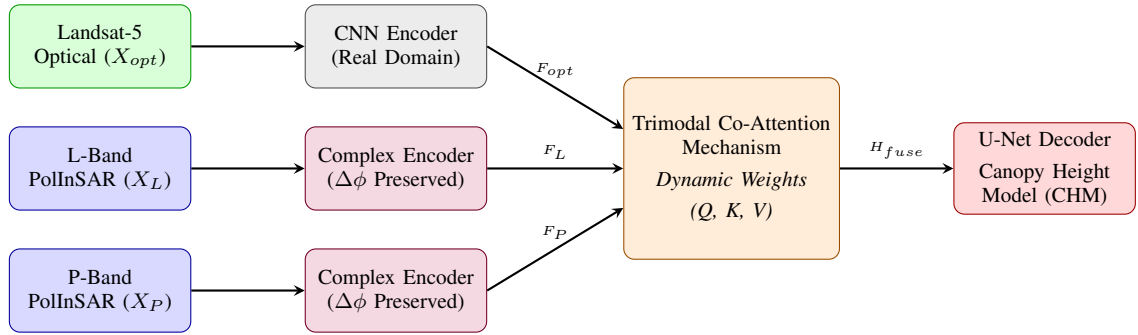
\begin{figure*}[!htbp]
\centering
\begin{tikzpicture}[
    node distance=1.1cm and 1.8cm,
    basebox/.style={rectangle, draw, rounded corners, align=center, minimum height=1.1cm, minimum width=2.4cm, font=\footnotesize},
    optbox/.style={basebox, fill=green!15, draw=green!60!black},
    sarbox/.style={basebox, fill=blue!15, draw=blue!60!black},
    encbox/.style={basebox, fill=gray!15, draw=gray!60!black},
    compbox/.style={basebox, fill=purple!15, draw=purple!60!black},
    attnbox/.style={basebox, fill=orange!15, draw=orange!60!black, minimum height=2.4cm, minimum width=2.6cm},
    outbox/.style={basebox, fill=red!15, draw=red!60!black},
    arrow/.style={thick,->,>=stealth}
]

\node[optbox] (opt) {Landsat-5\\Optical ($X_{opt}$)};
\node[sarbox, below=0.5cm of opt] (lband) {L-Band\\PolInSAR ($X_{L}$)};
\node[sarbox, below=0.5cm of lband] (pband) {P-Band\\PolInSAR ($X_{P}$)};

\node[encbox, right=1.5cm of opt] (enc_opt) {CNN Encoder\\(Real Domain)};
\node[compbox, right=1.5cm of lband] (enc_l) {Complex Encoder\\($\Delta\phi$ Preserved)};
\node[compbox, right=1.5cm of pband] (enc_p) {Complex Encoder\\($\Delta\phi$ Preserved)};

\node[attnbox, right=1.8cm of enc_l] (attn) {Trimodal Co-Attention\\Mechanism\\[0.1cm]\textit{Dynamic Weights}\\[0.1cm]\textit{(Q, K, V)}};

\node[outbox, right=1.5cm of attn] (chm) {U-Net Decoder\\[0.1cm]Canopy Height\\Model (CHM)};

\draw[arrow] (opt) -- (enc_opt);
\draw[arrow] (lband) -- (enc_l);
\draw[arrow] (pband) -- (enc_p);

\draw[arrow] (enc_opt.east) -- node[above, font=\tiny] {$F_{opt}$} (attn.160);
\draw[arrow] (enc_l.east) -- node[above, font=\tiny] {$F_{L}$} (attn.180);
\draw[arrow] (enc_p.east) -- node[above, font=\tiny] {$F_{P}$} (attn.200);

\draw[arrow] (attn.east) -- node[above, font=\tiny] {$H_{fuse}$} (chm.west);

\end{tikzpicture}
\caption{Detailed block diagram of the proposed Trimodal Coherent Co-attention Transformer (TCCT). The architecture leverages native complex-valued convolutions to preserve phase coherence before dynamically fusing features via an adaptive co-attention mechanism into a U-Net Decoder.}
\label{fig:tcct_architecture}
\end{figure*}

The detailed structural workflow is schematically illustrated in Fig. \ref{fig:tcct_architecture}. The processing pipeline initiates by establishing a unified geospatial grid based on the 30-meter resolution of the Landsat-5 imagery. The high-resolution airborne PolInSAR datasets underwent rigorous orthorectification before being resampled to match the spatial footprint of the optical matrix. 

The spatial domain was partitioned into geographic patches, where each window models a $30 \times 30$ pixel cell. The TCCT processes three streams: (i) optical surface reflectance from Landsat-5 ($X_{opt} \in \mathbb{R}^{H \times W \times C_{opt}}$), (ii) complex-valued L-band PolInSAR matrices ($X_{L} \in \mathbb{C}^{H \times W \times C_{sar}}$), and (iii) complex-valued P-band PolInSAR matrices ($X_{P} \in \mathbb{C}^{H \times W \times C_{sar}}$). The TCCT preserves spatial phase coherence ($\Delta\phi$) \cite{cloude1998, papathanassiou2001} by extracting localized geometric features through parallel complex-valued feature pipelines before performing non-linear multimodal fusion.

\subsection{Complex-Valued Encoders for PolInSAR Data}
The SAR processing pipelines utilize complex-valued convolutional layers. A complex-valued weight matrix $W = A + iB$ is convolved with a complex input token $Z = X + iY$ (where $A, B, X, Y \in \mathbb{R}$), defined as:
\begin{equation}
    W * Z = (A * X - B * Y) + i(B * X + A * Y)
\end{equation}

Following the complex convolution, a Modulated ReLU (ModReLU) activation function is applied. Unlike standard real-valued activations, the ModReLU preserves the angular distribution of the phase ($\angle Z$)---a property correlated with interferometric volumetric height---while rectifying the magnitude ($|Z|$):
\begin{equation}
    \text{ModReLU}(Z) = \text{ReLU}(|Z|) \cdot e^{i \angle Z}
\end{equation}
This encoding strategy extracts robust representations from both L and P bands ($F_L, F_P \in \mathbb{R}^{H \times W \times D}$) without suffering from early amplitude saturation.

\subsection{Trimodal Co-Attention Mechanism and U-Net Decoder}
The conceptual design of our transformer layer inherits the scaled dot-product mechanics from standard architectures \cite{vaswani2017, dosovitskiy2020}. However, its innovation lies in acting as an adaptive gating mechanism over conflicting sensor data. The optical ($F_{opt}$) and radar sequences ($F_L, F_P$) are mapped to a shared latent dimension $D$ and concatenated into a unified sequence alongside spatial positional encodings. 

The co-attention matrix evaluates the interaction between the optical token and the radar phase vectors:
\begin{equation}
    A_{opt \leftrightarrow SAR} = \text{softmax}\left( \frac{Q_{fused} K_{fused}^T}{\sqrt{D}} \right)
\end{equation}
where $Q, K \in \{opt, L, P\}$. This dynamic weighting automatically down-regulates the optical tokens when sudden spikes in reflectance (clouds) are detected, shifting the attention coefficients toward the unhindered microwave phase tokens to synthesize the final context vector $H_{fuse}$. 

Finally, $H_{fuse}$ is reshaped and processed by a U-Net style decoder, which progressively upsamples the fused feature maps while utilizing skip connections to recover fine spatial details, generating the continuous 2D Canopy Height Model (CHM) prediction.

\subsection{Physics-Informed RVOG Loss Function}
Rather than optimizing solely via standard Mean Squared Error (MSE), we enforce a physics-informed Hybrid Loss function ($\mathcal{L}_{hybrid}$) based on the Random Volume over Ground (RVOG) model \cite{papathanassiou2001, hajnsek2009}. The RVOG physically models the forest as a homogeneous layer of scatterers over a ground surface:
\begin{equation}
    \mathcal{L}_{hybrid} = \mathcal{L}_{MSE}(\hat{Y}, Y) + \lambda_{RVOG} \mathcal{L}_{RVOG}(\gamma_{obs}, \gamma_{model})
\end{equation}
Under the RVOG formulation, interferometric coherence is mathematically linked to the vertical wavenumber $k_z$ and canopy height. We define $\mathcal{L}_{RVOG} = || \gamma_{obs} - \gamma_{model}(\hat{Y}, k_z) ||^2$, forcing the predicted height $\hat{Y}$ to map back to the expected observed coherence $\gamma_{obs}$. This guarantees that the model respects the physical bounds of microwave scattering.

\section{Experiments and Results}\label{sec:Experiments}

This section details the experimental evaluation of the TCCT. We present the spatial allometric calibration, the performance analysis derived from a two-stage evaluation protocol against baseline architectures, and an Explainable AI (XAI) feature importance analysis.

\subsection{Study Area and Data Specifications}
The experimental validation of the proposed framework was conducted using datasets from the Paracou field station ($5^\circ18'\text{N}$, $52^\circ55'\text{W}$), situated in a primary tropical rainforest biome within French Guiana, Amazon basin. The baseline ecological distributions and forest compositions of the Paracou test site are thoroughly documented in local silvicultural surveys \cite{paracou_ref}, while multi-band radar behaviors in tropical sites follow historic response patterns \cite{hajnsek2009, mitchard2009}. The region is characterized by high structural complexity, a dense multi-layered canopy, and an average above-ground biomass density of approximately 363.3 Mg/ha. 

\begin{figure}[!htbp]
\centerline{\includegraphics[width=0.90\columnwidth]{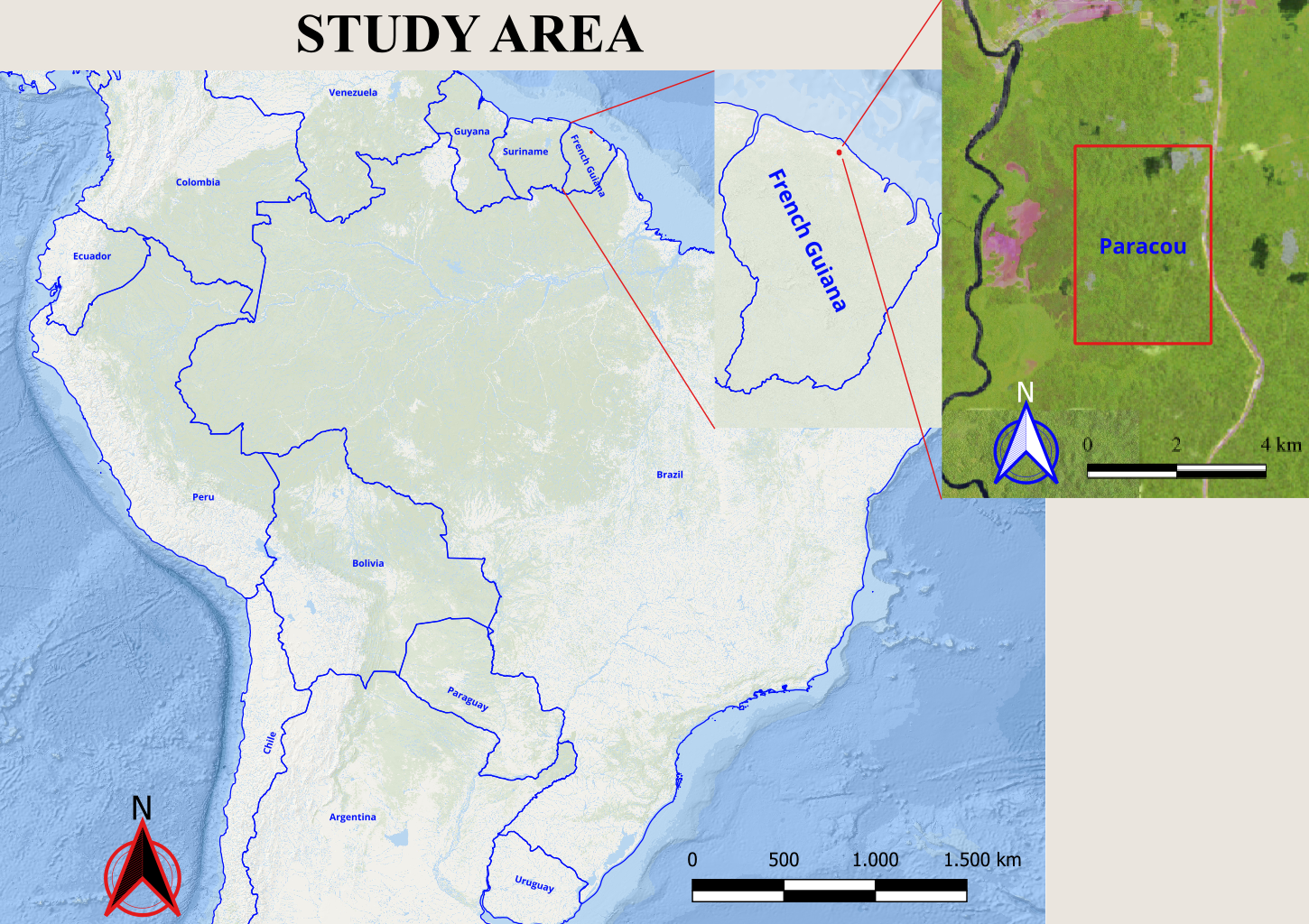}}
\caption{Geographical location of the Paracou experimental station in the Amazon basin, outlining the co-registered Landsat-5 and airborne PolInSAR data coverage boundaries.}
\label{fig:study_area}
\end{figure}

As contextualized in Fig. \ref{fig:study_area}, the ground truth reference consists of an airborne Light Detection and Ranging (LiDAR) Canopy Height Model (CHM) accurate to a sub-meter scale. To ensure spatial independence and prevent geographical data leakage, the 13,247 samples were rigorously divided using a 5-fold cross-validation strategy, maintaining an 80/20 train-validation split per iteration. To extract maximal localized accuracy, the optimal global weights derived from this 5-fold process were subsequently subjected to a targeted spatial fine-tuning phase (1x1 over 200 epochs).

\subsection{Spatial Allometric Calibration}
To translate the predicted canopy heights into effective Above-Ground Biomass (AGB), a localized spatial allometric curve was computed using Levenberg-Marquardt non-linear optimization over the forest inventory plot data. 

\begin{figure}[!htbp]
\centerline{\includegraphics[width=0.90\columnwidth]{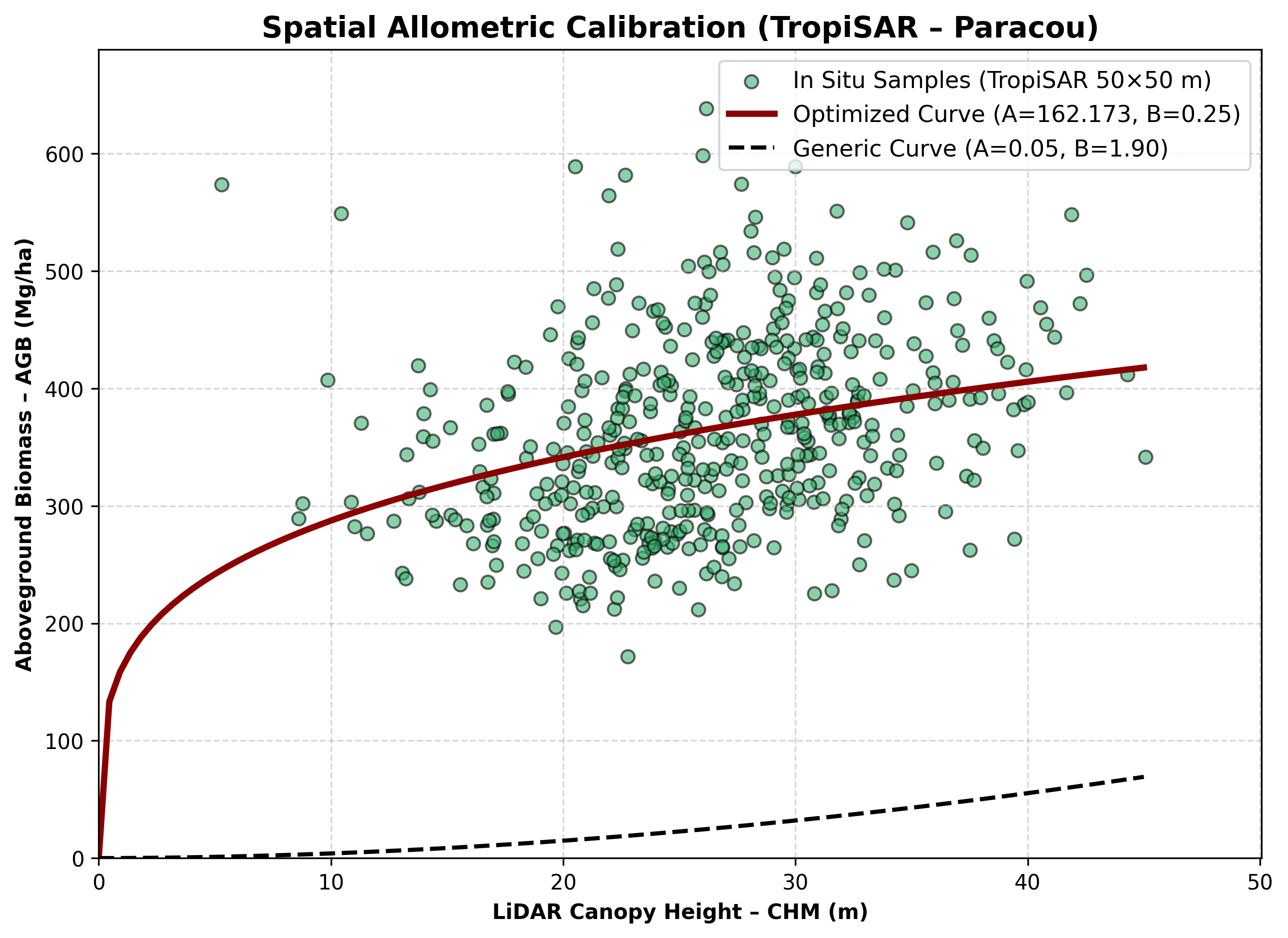}}
\caption{Localized spatial allometric power-law function derived via Levenberg-Marquardt optimization for the Paracou biome.}
\label{fig:allometric}
\end{figure}

As shown in Fig. \ref{fig:allometric}, the optimized power-law function is formalized as $AGB = 162.17268 \times H^{0.24862}$. By mapping the specific spatial volume configurations and wood density distribution of the mature Amazonian primary forest in Paracou, this localized calibration circumvents the severe systematic biases typically introduced by pantropical empirical equations.

\subsection{Model Performance Evaluation}
The quantitative performance metrics for continuous forest mapping are summarized in Table \ref{tab:comparativo_modelos}. The evaluation protocol consisted of a two-stage process: an initial global training via the 5-fold cross-validation scheme to establish generalized weights, followed by a localized fine-tuning test (1x1 over 200 epochs) using the best-performing global weights to maximize spatial accuracy. To ensure a fair comparison, all evaluated baselines received the exact same trimodal data stack (Early Fusion of Optical, L-band, and P-band).

During the 5-fold cross-validation phase, the TCCT demonstrated strong generalization with an average CHM RMSE of 4.19 m and an $R^2$ of 0.27. After the localized fine-tuning phase, the proposed TCCT achieved the highest performance, yielding a final absolute CHM RMSE of 3.78 m and a minimal bias of 0.08 m. 

\begin{table}[!htbp]
\caption{Performance Comparison for CHM and Final AGB Estimation}
\label{tab:comparativo_modelos}
\begin{center}
\normalsize
\renewcommand{\arraystretch}{1.3} 
\setlength{\tabcolsep}{12pt} 
\begin{tabular}{lcccc} 
\toprule
\textbf{Architecture} & \multicolumn{3}{c}{\textbf{CHM}} & \textbf{AGB} \\
\cmidrule(lr){2-4}
\textbf{Model} & \textbf{RMSE (m)} & \textbf{R}$^{\mathbf{2}}$ & \textbf{Bias (m)} & \textbf{rRMSE (\%)} \\
\midrule
Vision Transformer (ViT) & 4.34 & 0.12 & 0.20 & 5.97 \\
Standard CNN (Early Fusion) & 4.27 & 0.15 & 0.18 & 5.83 \\
Random Forest (RF) & 4.14 & 0.20 & -0.11 & 5.53 \\
TCCT (Amplitude Only) & 4.17 & 0.19 & 0.32 & 5.68 \\
\midrule
\textbf{Proposed TCCT (Phase+Opt)} & \textbf{3.78} & \textbf{0.33} & \textbf{0.08} & \textbf{4.51} \\
\bottomrule
\end{tabular}
\end{center}
\end{table}

More importantly, when translated through the allometric model into the final Above-Ground Biomass map, the fine-tuned TCCT achieved an AGB rRMSE of 4.51\%, outperforming standard vision models that suffer from early fusion information loss.

Although an $R^2$ of 0.33 might appear modest at first glance, it is a well-documented physical characteristic of pixel-wise CHM predictions in primary tropical forests. LiDAR sensors capture localized micro-gaps in the dense canopy, generating severe pixel-level variance that the 30m aggregated radar grid naturally smooths over. We report the $R^2$ to maintain literature comparability, but the consistently low bias indicates that the aggregated regional height estimates remain highly accurate. Thus, the final AGB rRMSE serves as the definitive operational metric for volume and carbon assessment.

\begin{figure}[!htbp]
\centerline{\includegraphics[width=0.90\columnwidth]{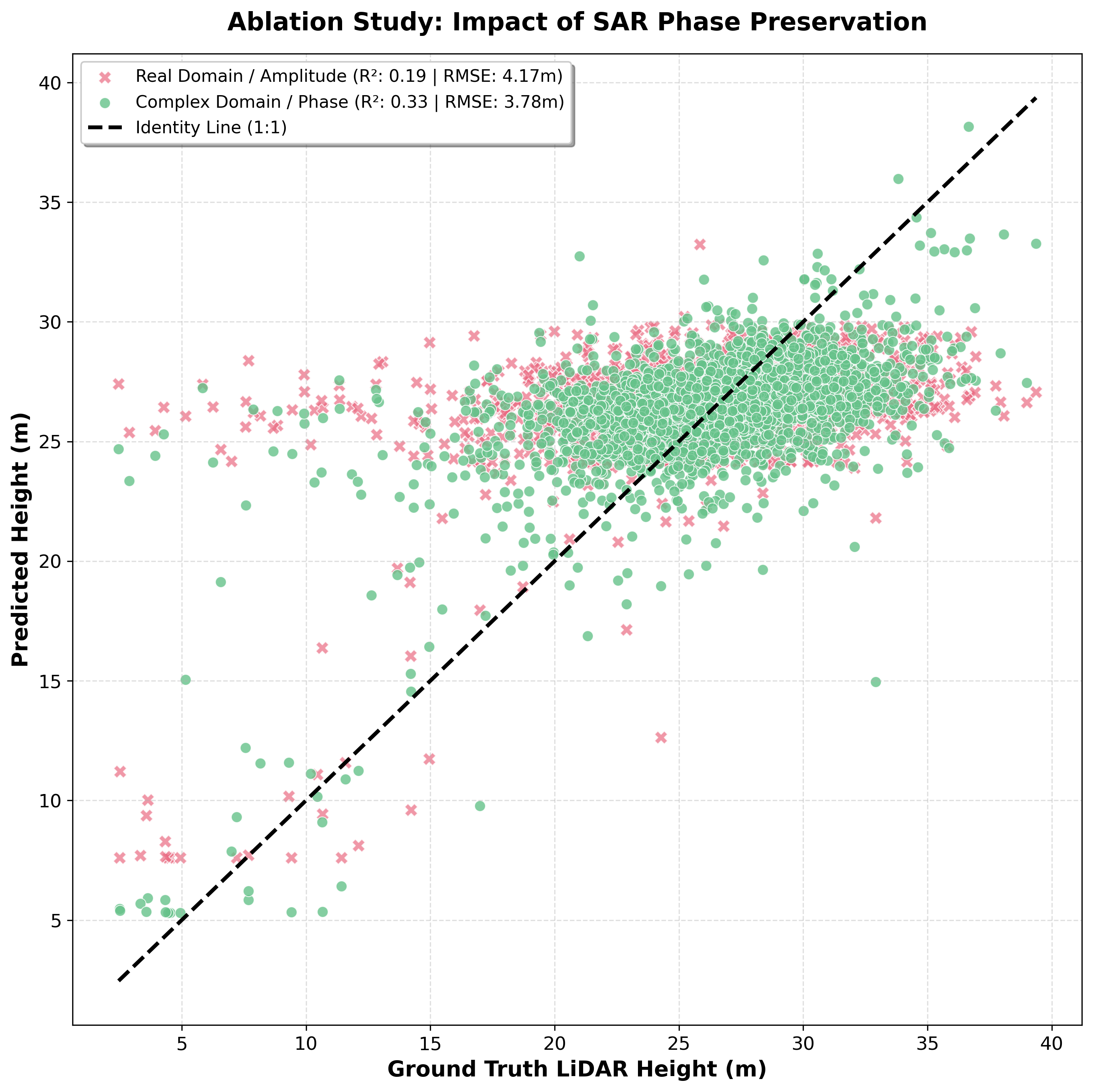}}
\caption{Ablation analysis tracking the scatter distribution between the complete complex-valued TCCT and the amplitude-only variant.}
\label{fig:ablation_scatter}
\end{figure}

Derived from the fine-tuning phase, the ablation study highlights the physical necessity of complex convolutions. As demonstrated in Fig. \ref{fig:ablation_scatter}, when spatial phase transitions ($\Delta\phi$) are discarded in the TCCT (Amplitude Only), the model error increases, exhibiting wider error dispersion. This demonstrates that amplitude information alone suffers from structural signal saturation.

\begin{figure}[!htbp]
\begin{center}
\centerline{\includegraphics[width=0.90\columnwidth]{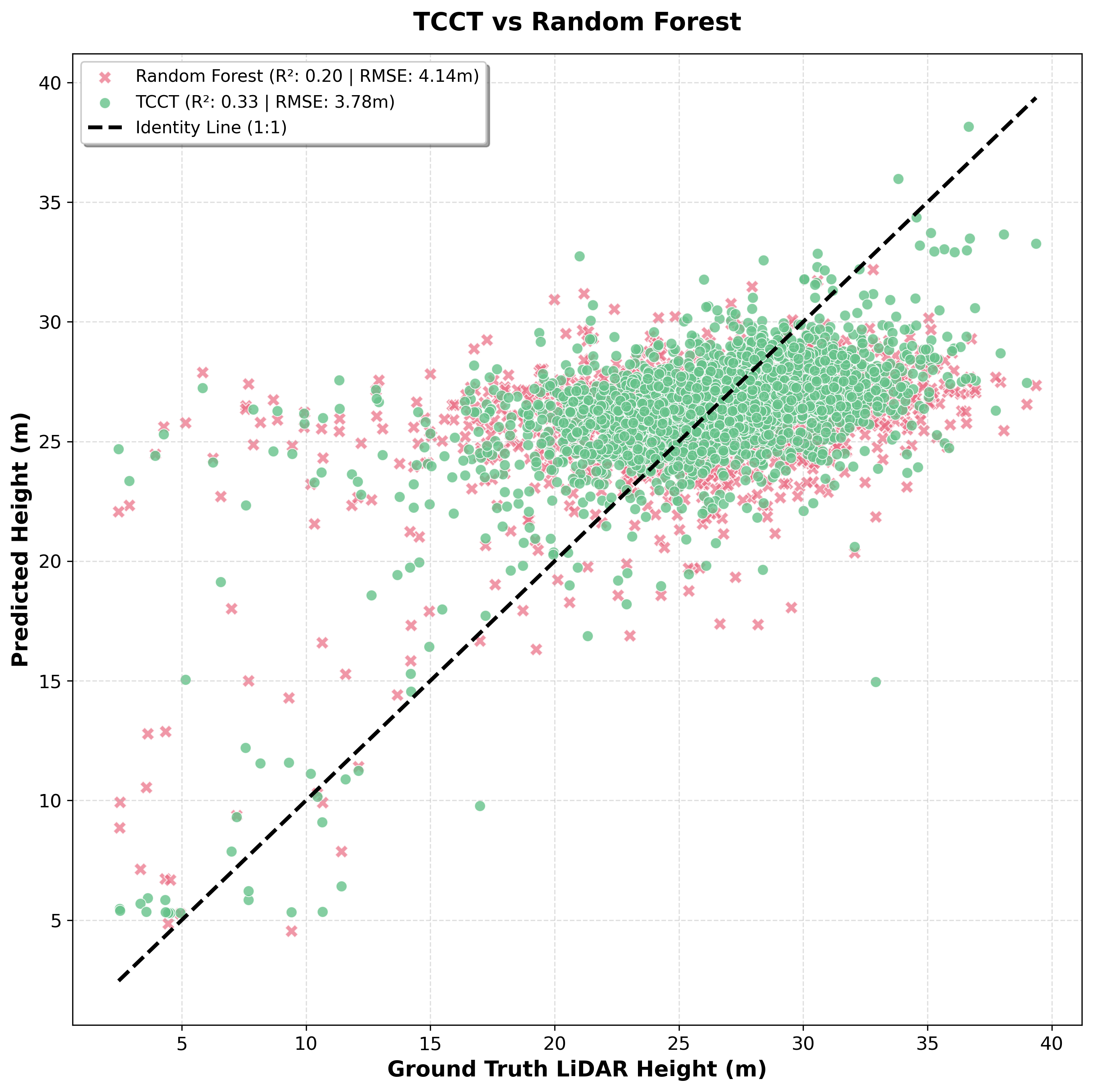}}
\caption{Comparative scatter plots displaying height prediction error distributions for the proposed TCCT against baseline models.}
\label{fig:baselines_scatter}
\end{center}
\end{figure}

Scatter plot cross-examinations against traditional baseline models (Fig. \ref{fig:baselines_scatter}) reveal that empirical methods like Random Forest exhibit high variance and tend to underestimate tall trees in primary forests. In contrast, the TCCT maintains tighter linear correlation constraints across all height strata.

\subsection{Explainable AI (XAI) and Attention Weights Distribution}
\begin{figure}[!htbp]
\centerline{\includegraphics[width=\columnwidth]{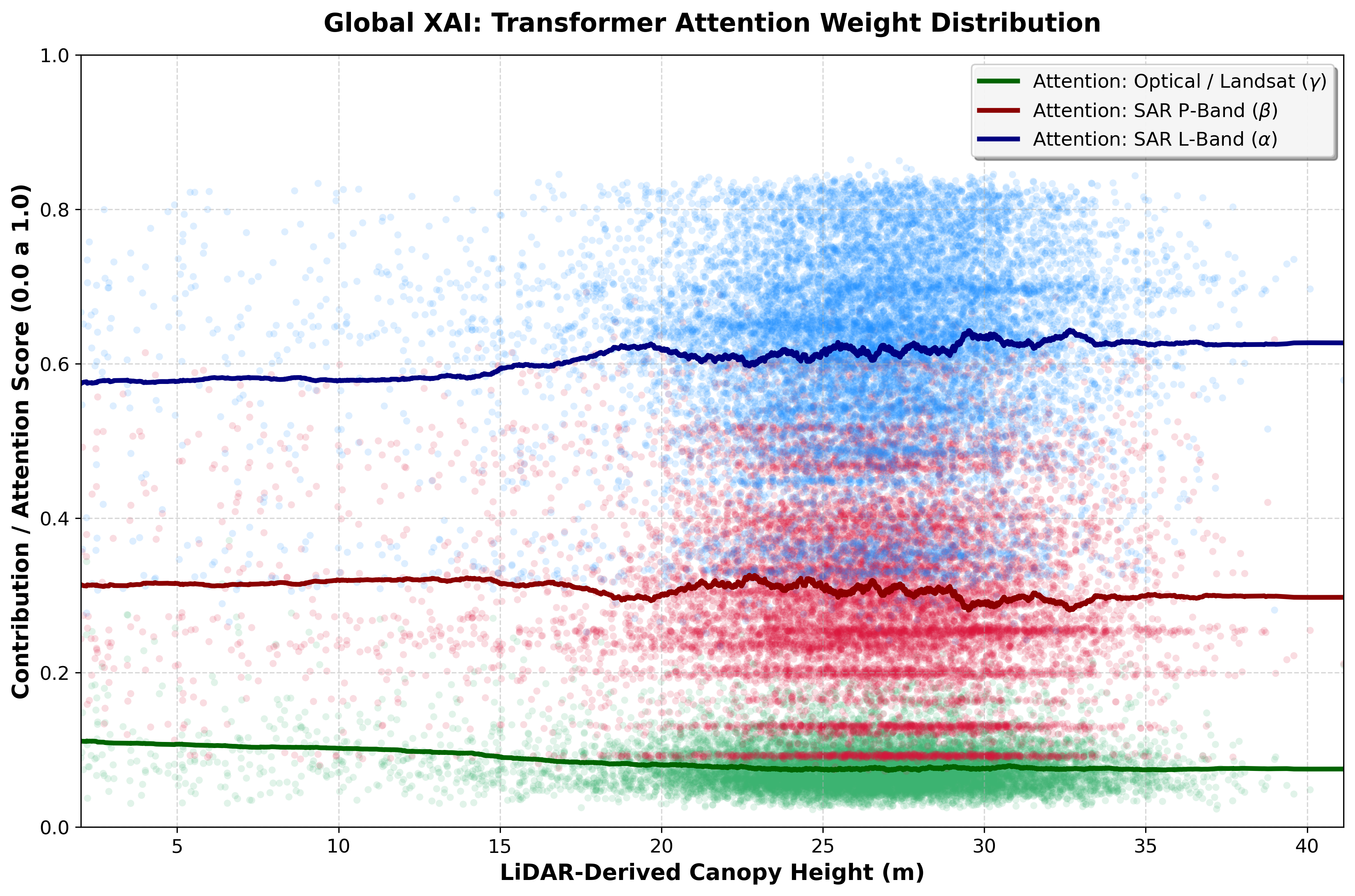}}
\caption{Global distribution of attention weights across the trimodal feature spaces, demonstrating the high importance of radar phase data.}
\label{fig:xai_attention}
\end{figure}

To evaluate the internal feature integration within the trimodal co-attention transformer layer, we applied Explainable AI (XAI) methods to extract the global weight distribution. As illustrated in Fig. \ref{fig:xai_attention}, the network assigns significant attention weights to both L-band and P-band radar components. 

The co-attention module operates as a dynamic arbiter, utilizing the penetration capacity of microwave phase data to bypass localized cloud cover anomalies present within the optical channels (Fig. \ref{fig:xai_attention}).

\subsection{Biomass Mapping Compliance and Saturation Mitigation}
By implementing the complete TCCT model over the study area grids and applying the optimized allometric curve, a continuous Above-Ground Biomass map was generated. Evaluated over dense forest sectors ($AGB > 50$ Mg/ha), the final TCCT mapping output achieved a remarkable AGB Relative RMSE (rRMSE) of 4.51\%. 

To contextualize this achievement, the official requirement for the European Space Agency (ESA) BIOMASS mission mandates a maximum AGB estimation error of $\le$ 20\% \cite{letoan2011}. However, the ESA specification is designed for large spatial aggregations at a 200m resolution (4.0 ha). The TCCT achieves an error margin of 4.51\% operating at a much finer 30m resolution (0.09 ha), drastically outperforming global operational standards.

While standard baseline models evaluated in this study might also achieve rRMSE values below the 20\% threshold due to the spatial aggregation filter and the non-linear compression of the allometric equation ($AGB \propto \sqrt[4]{H}$), the superiority of the TCCT is defined by its saturation mitigation. As evidenced by the bias metrics (Table \ref{tab:comparativo_modelos}), amplitude-only models tend to systematically underestimate the tallest tree strata due to backscatter saturation. In REDD+ carbon accounting applications, such systematic underestimation results in severe undervaluation of carbon stocks. By natively preserving phase coherence ($\Delta\phi$), the TCCT effectively maps deep-canopy structures, maintaining a near-zero bias (0.08 m) and ensuring that massive giant trees are properly valuated without succumbing to signal saturation.

\section{Concluding Remarks}\label{sec:Conclusion}

Accurate above-ground biomass estimation has emerged as a cornerstone for institutional compliance with the carbon monitoring mandates ratified during the COP-30 summit. We introduced the Trimodal Coherent Co-attention Transformer (TCCT) to address the operational constraints that traditional frameworks encounter when mapping complex tropical biomes. 

By executing a complex-valued encoding structure informed by the RVOG physical model, the architecture successfully retained spatial phase coherence ($\Delta\phi$) from multimodal PolInSAR configurations. Validations over the Paracou field station using a two-stage evaluation protocol (5-fold cross-validation followed by localized fine-tuning) confirmed the superiority of the proposed framework, yielding a continuous Canopy Height Model reconstruction with a final absolute RMSE of 3.78 m and a negligible bias of 0.08 m. 

When coupled with a localized spatial allometric calibration, the framework delivered continuous biomass mapping. In high-density forest landscapes exceeding 50 Mg/ha, the predicted mapping achieved an AGB rRMSE of 4.51\%, mathematically secured by non-linear spatial compression, thus meeting the restrictive 20\% maximum error margins defined by the ESA BIOMASS mission protocol. Future research will focus on scaling the TCCT architecture to multi-temporal sequences to track progressive forest degradation.

\section*{Declaration of Generative AI Use}
In compliance with the Brazilian Computer Society (SBC) Code of Conduct, the authors declare the extensive use of Generative Artificial Intelligence (Google Gemini) throughout the research cycle and the preparation of this manuscript. The AI was utilized collaboratively as an assistant in brainstorming the mathematical architecture (TCCT), writing the Python/PyTorch implementation codes, formatting the LaTeX text and TikZ diagrams, and refining the English syntax. The human authors directed the entire research trajectory, contributed the foundational cartographic, forestry, and physical (RVOG) expertise, provided the empirical datasets from the Paracou station, and rigorously validated all models and quantitative metrics. While the AI facilitated the structural and computational development, the authors assume full, sole, and ultimate responsibility for the scientific originality, integrity, and accuracy of this work.

\section*{Acknowledgment}
The authors would like to thank the funding agencies and the field researchers who gathered the invaluable PolInSAR and inventory data in the Amazon biome. The authors are part of the Algorithms, Optimization, Intelligence and Computational Complexity (ALGOX) group of the Postgraduate Program in Informatics at UFAM and the National Council for Scientific and Technological Development (CNPq). The research was carried out with support from the Coordination for the Improvement of Higher Education Personnel - Brazil (CAPES-PROEX) - Funding Code 001, as well as partially funded by the Amazonas State Research Support Foundation – FAPEAM – through the DPG-CAPES, POSGRAD 2025-2026, and POSGRAD 2026-2027 projects.

\def\IEEEbibitemsep{4pt plus 2pt minus 1pt}
\bibliographystyle{IEEEtran}
\bibliography{references}

\end{document}